\documentclass[a4paper, 10pt, conference]{ieeeconf}      
\usepackage{FG2023}

\FGfinalcopy 

\IEEEoverridecommandlockouts                              
\usepackage{graphics} 
\usepackage{epsfig} 
\usepackage{mathptmx} 
\usepackage{times} 
\usepackage{amsmath} 
\usepackage{amssymb}  
\usepackage{multirow}
\usepackage{subcaption}
\usepackage{caption}
\usepackage{booktabs}
\usepackage[mathscr]{euscript}

\def\FGPaperID{0001} 

\title{\LARGE \bf Localization using Multi-Focal Spatial Attention \\ for Masked Face Recognition
}

\author{\parbox{16cm}{\centering
    {\large Yooshin Cho$^{1}$ ~~~ Hanbyel Cho$^{1}$ ~~~ Hyeong Gwon Hong$^{2}$ ~~~ Jaesung Ahn$^{2}$ \\ \vspace{1mm} Dongmin Cho$^{3}$ ~~~ JungWoo Chang$^{3}$ ~~~ Junmo Kim$^{1,2}$}\\
    \vspace{1mm}
    {\normalsize
    $^{1}$School of Electrical Engineering, KAIST, South Korea \\
$^{2}$Kim Jaechul Graduate School of AI, KAIST, South Korea} \\
$^{3}$Alchera Inc}
    \thanks{This work was not supported by any organization}
}

\begin{document}

\ifFGfinal
\thispagestyle{empty}
\pagestyle{empty}
\else
\author{Anonymous FG2023 submission\\ Paper ID \FGPaperID \\}
\pagestyle{plain}
\fi
\maketitle

\begin{abstract}

Since the beginning of world-wide COVID-19 pandemic, facial masks have been recommended to limit the spread of the disease. However, these masks hide certain facial attributes. Hence, it has become difficult for existing face recognition systems to perform identity verification on masked faces. In this context, it is necessary to develop masked Face Recognition (MFR) for contactless biometric recognition systems. Thus, in this paper, we propose Complementary Attention Learning and Multi-Focal Spatial Attention that precisely removes masked region by training complementary spatial attention to focus on two distinct regions: masked regions and backgrounds. In our method, standard spatial attention and networks focus on unmasked regions, and extract mask-invariant features while minimizing the loss of the conventional Face Recognition (FR) performance. For conventional FR, we evaluate the performance on the IJB-C, Age-DB, CALFW, and CPLFW datasets. We evaluate the MFR performance on the ICCV2021-MFR/Insightface track, and demonstrate the improved performance on the both MFR and FR datasets. Additionally, we empirically verify that spatial attention of proposed method is more precisely activated in unmasked regions. 

\end{abstract}

\section{INTRODUCTION}

With the advent of deep neural networks, the accuracy of Face Recognition (FR) has become more than over 99\% in controlled environments~\cite{deng2019arcface, wang2018cosface, boutros2022elasticface}. Accordingly, identity authentication systems that employ FR have been widely used in our daily life (e.g., airports, companies, and smartphones). However, owing to the recent global COVID-19 pandemic, it has become mandatory to wear facial masks to protect public health. These facial masks cover the nose and mouth, and hence, they significantly decrease the performance of previous face recognition systems. Thus, the necessity of Masked Face Recognition (MFR) has been highlighted, and various MFR studies have been proposed during the pandemic.

Previous studies on MFR have focused on the effective extraction of mask-invariant features~\cite{zhang2022learning, hemathilaka2022comprehensive,qi2021balanced, neto2021focusface, deng2021masked}. The method proposed in~\cite{zhang2022learning} discards the features obtained from the lower half region of images by utilizing the prior knowledge that a mask is located in the lower half of a face. In addition, MMD loss, MSE loss, or adversarial loss with an auxiliary mask-usage classification branch has been adopted to reduce the distance between the features obtained from masked and unmasked images~\cite{neto2021focusface, huber2021mask}. These approaches have successfully improved the MFR performance by forcing networks to neglect the facial attributes (nose and mouth) that are typically behind a mask. However, this conflicts with the goal of conventional FR and leads to performance degradation on FR datasets.

\begin{figure}[t]
    \centering
    \begin{subfigure}[b]{0.09\textwidth}
        \centering
        \includegraphics[width=\textwidth]{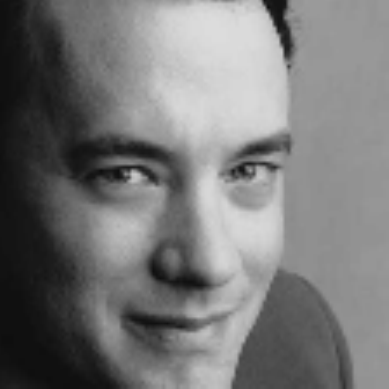}
        \subcaption{Original}
        \label{subfig:org}
        
    \end{subfigure}
    \hfill
    \begin{subfigure}[b]{0.09\textwidth}
        \centering
        \includegraphics[width=\textwidth]{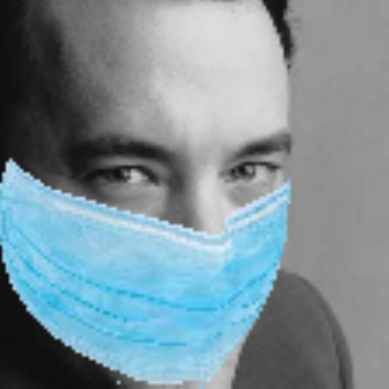}
        \subcaption{Blue}
        \label{subfig:blue}
        
    \end{subfigure} 
    \hfill
    \begin{subfigure}[b]{0.09\textwidth}
        \centering
        \includegraphics[width=\textwidth]{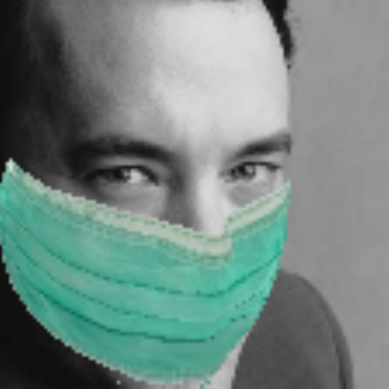}
        \subcaption{Green}
        \label{subfig:green}
        
    \end{subfigure} 
    \hfill
    \begin{subfigure}[b]{0.09\textwidth}
        \centering
        \includegraphics[width=\textwidth]{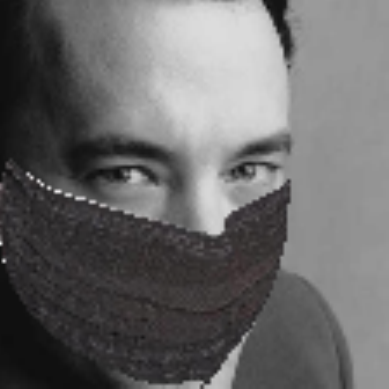}
        \subcaption{Black}
        \label{subfig:black}
    \end{subfigure}
    \hfill    
    \begin{subfigure}[b]{0.09\textwidth}
        \centering
        \includegraphics[width=\textwidth]{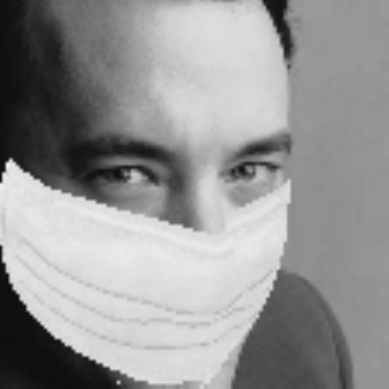}
        \subcaption{White}
        \label{subfig:white}
    \end{subfigure} 
    \caption{Examples of the synthetic masked face, which is generated by the online masked face generation function of the ICCV2021/Insightface track.}
    \vspace{-6mm}
    \label{fig:syn_masks}
\end{figure}

To optimize the trade-off between the performances of conventional FR and MFR, we propose the method that more precisely separates masked regions from unmasked region by adopting a spatial attention module~\cite{cho2022rethinking}. Conventionally, spatial attention modules, such as the Convolutional Block Attention Module (CBAM)~\cite{woo2018cbam} have been adopted to make networks focus on foreground objects, and have demonstrated the localization capability that separates foreground regions from background regions. In this paper, we propose Complementary Attention Learning (CAL) that adversarially utilizes the complementary spatial attention to enhance the localization capability of CBAM inspired by recent studies on unbiased visual recognition~\cite{wang2021causal}. We train complementary spatial attention to learn undesirable information (e.g., mask-usage classification) to prevent standard spatial attention from focusing on the undesirable region (e.g., masked region). We describe details in Section~\ref{subsec:comple}

Additionally, we propose Multi-Focal Spatial Attention (MFSA), which separates an image into three regions (i.e., unmasked regions, masked regions, and background regions). Previously, $sigmoid$ function has been generally adopted to normalize spatial attention of CBAM, and divided the elements into binary (e.g., foreground and background). Thus, if we train complementary attention to focus on masked region, background region is more likely to be activated at standard attention. To alleviate the issue, we employ $softmax$ function to classify elements into N-way. Also, instead of channel attention module of CBAM, we employ the convolution layers that followed by batch normalization and ReLU. To disentangle the representation of N-way attention, we adopt the Orthogonalization by Newton's Iteration (ONI)~\cite{huang2020controllable}, and it orthogonalizes the weight of convolution layers. Weight orthogonalization enforces each output channel to depend on a different channel of input channel, and it reduces causality and correlation between output channels. With the modification, we increase the model representation capacity, and empirically demonstrate improved performance on both MFR and FR datasets.

\begin{figure}[t]
    \centering
    \begin{subfigure}[b]{0.48\textwidth}
        \centering
        \includegraphics[width=\textwidth]{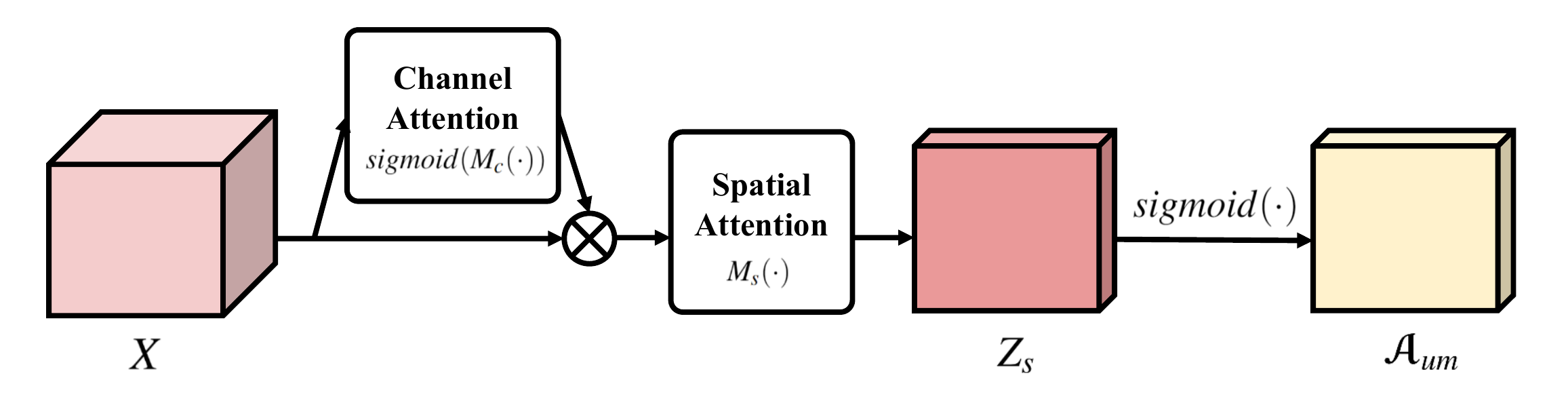}
         \subcaption{Convolutional Block Attention Module (CBMA)}
         \label{subfig:cbam}
    \end{subfigure} 
    \vspace{2mm}
    \begin{subfigure}[b]{0.48\textwidth}
        \centering
        \includegraphics[width=\textwidth]{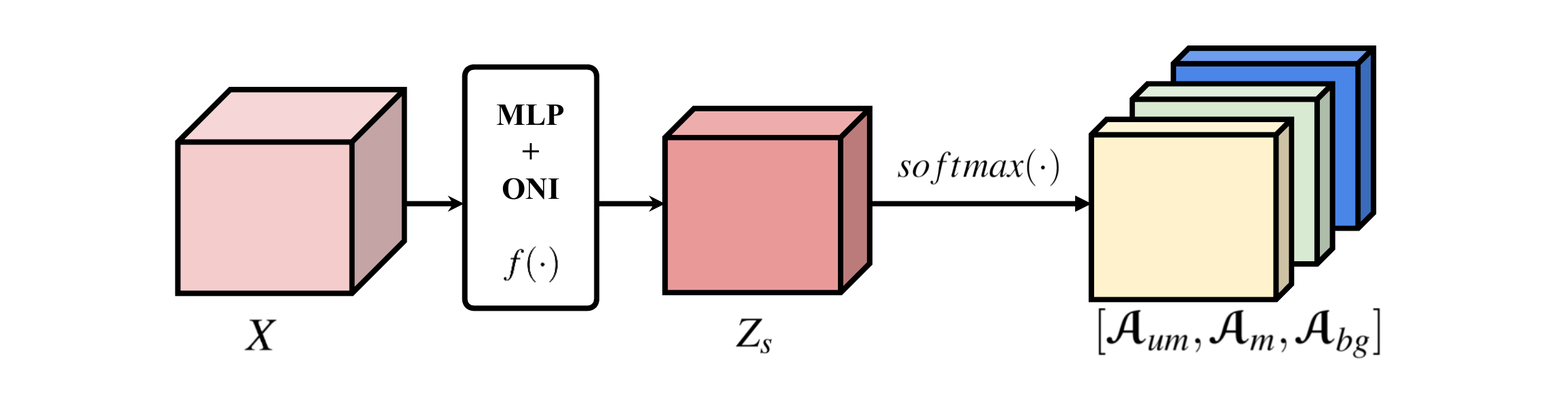}
        \subcaption{Multi-Focal Spatial Attention (MFSA)}
        \label{subfig:ours}

    \end{subfigure}
    \caption{\textbf{Schematic of the Convolutional Block Attention Module (CBAM) and Multi-Focal Spatial Attention (MFSA).} We replace the channel attention of CBAM and the last $sigmoid$ function with the convolutional layers and $softmax$ function, respectively. Additionally, we adopt the Orthogonalization by Newton's Iteration (ONI) to the weight of convolutional layers to disentangle the representations.}
    \vspace{-2mm}
    \label{fig:att_modules}
\end{figure}

We follow the experimental setup of the baseline models presented at the Insightface track of the ICCV2021-MFR challenge~\cite{deng2021masked}. We train networks on the CASIA-Webface~\cite{yi2014learning} and MS1MV3 datasets~\cite{deng2019arcface, deng2020retinaface, guo2016ms} with masked face augmentation that synthesizes masked face with a given probability. To implement masked face augmentation, we employ the online masked face generation function provided by Insightface. We visualize the examples of synthetic masked face in the Fig~\ref{fig:syn_masks}. We train networks with randomly synthesized masked face, and evaluate on the ICCV2021-MFR/Insightface track, which is a \textbf{real masked face} dataset. To analyze the trade-off between MFR and FR performances, we evaluate the FR performance on IJB-C, Age-DB, CALFW, and CPLFW datasets~\cite{CPLFWTech, moschoglou2017agedb,DBLP:journals/corr/abs-1708-08197, maze2018iarpa}. From the results, we empirically verify that proposed method achieves the superior performance on both FR and MFR datasets.

As wearing masks have been highly recommended, many disciplines that require to recognize and interact with human need to develop masked face recognition algorithms (e.g., cyber-security, transportation, public health, human-computer interaction, and smart technologies). Especially, our method demonstrates the masked region by spatial attention maps, so our method is explainable to human, which is an essential behavior for human-computer interaction. Also, our method can be applied to other type of facial occlusions (e.g., glasses and hats), if synthetic image generator is exist. Therefore, we expect our algorithm can be ubiquitously applied to other disciplines.

\section{Localization using Multi-Focal Spatial Attention}

We propose the method that precisely localizes the unmasked region of a face. First, we propose the Complementary Attention Learning (CAL), which prevents spatial attention from being activated in an undesirable area, such as the masked region of a face. Second, we propose the Multi-Focal Spatial Attention (MFSA), which does not only divide the region into binary (e.g., foreground and background), but divides into 3-way (e.g., masked region, unmasked region, and background region). Details of the proposed method are described in the following sections.

\subsection{Preliminary: CBAM}
\label{subsec:CBAM}
Before introducing our method, we briefly describe Convolutional Block Attention Module (CBAM)~\cite{woo2018cbam}. CBAM is a representative attention module that sequentially applies channel attention and spatial attention modules. It can be expressed using the following formulas:
\begin{align}
\label{eq:cbam}
    \boldsymbol{\mathscr{A}_{c}} &= sigmoid ( \boldsymbol{M_c(X)} ),& \boldsymbol{X_c} &=  \boldsymbol{\mathscr{A}_{c}}  \otimes \boldsymbol{X}, &  \\
    \boldsymbol{\mathscr{A}_{s}} &= sigmoid ( \boldsymbol{M_s(X_c)} ),& \boldsymbol{Y} &= \boldsymbol{\mathscr{A}_{s}} \otimes  \boldsymbol{X_c},
\end{align}
where $\boldsymbol{X}, \boldsymbol{Y} \in \mathbb{R}^{C \times H \times W}$ are the input and output features of the CBAM, respectively. $\otimes$ denotes element-wise multiplication, and $\boldsymbol{M_c(\cdot)$ and $M_s(\cdot)}$ are the channel and spatial attention modules, respectively. $X_c \in \mathbb{R}^{C \times H \times W}$ is an intermediate feature refined with channel attention. The channel attention module $\boldsymbol{M_c(\cdot)}$ is composed of max-pooling, average-pooling along spatial dimensions, and multi-layer perceptrons (MLPs). The spatial attention module $\boldsymbol{M_s(\cdot)}$ is composed of max-pooling, average-pooling along a channel dimension, and convolution layers. $\boldsymbol{M_c(\boldsymbol{X})} \in \mathbb{R}^{C \times 1 \times 1}$, $\boldsymbol{M_s(\boldsymbol{X_c})} \in \mathbb{R}^{1 \times H \times W}$ are normalized by $sigmoid$ function to compute the channel and spatial attention, respectively. Channel attention $\boldsymbol{\mathscr{A}_{c}} \in \mathbb{R}^{C \times 1 \times 1}$ and spatial attention $\boldsymbol{\mathscr{A}_{s}} \in \mathbb{R}^{1 \times H \times W}$ represent the importance of channel and spatial locations, respectively. Empirically, the CBAM demonstrates the localization capability, and $\boldsymbol{\mathscr{A}_{s}}$ focuses on the foreground objects. In this work, we enhance the localization capability to make the network precisely focus on the unmasked region of a face.

\begin{figure*}[!t]
    \centering
    \includegraphics[width=\linewidth]{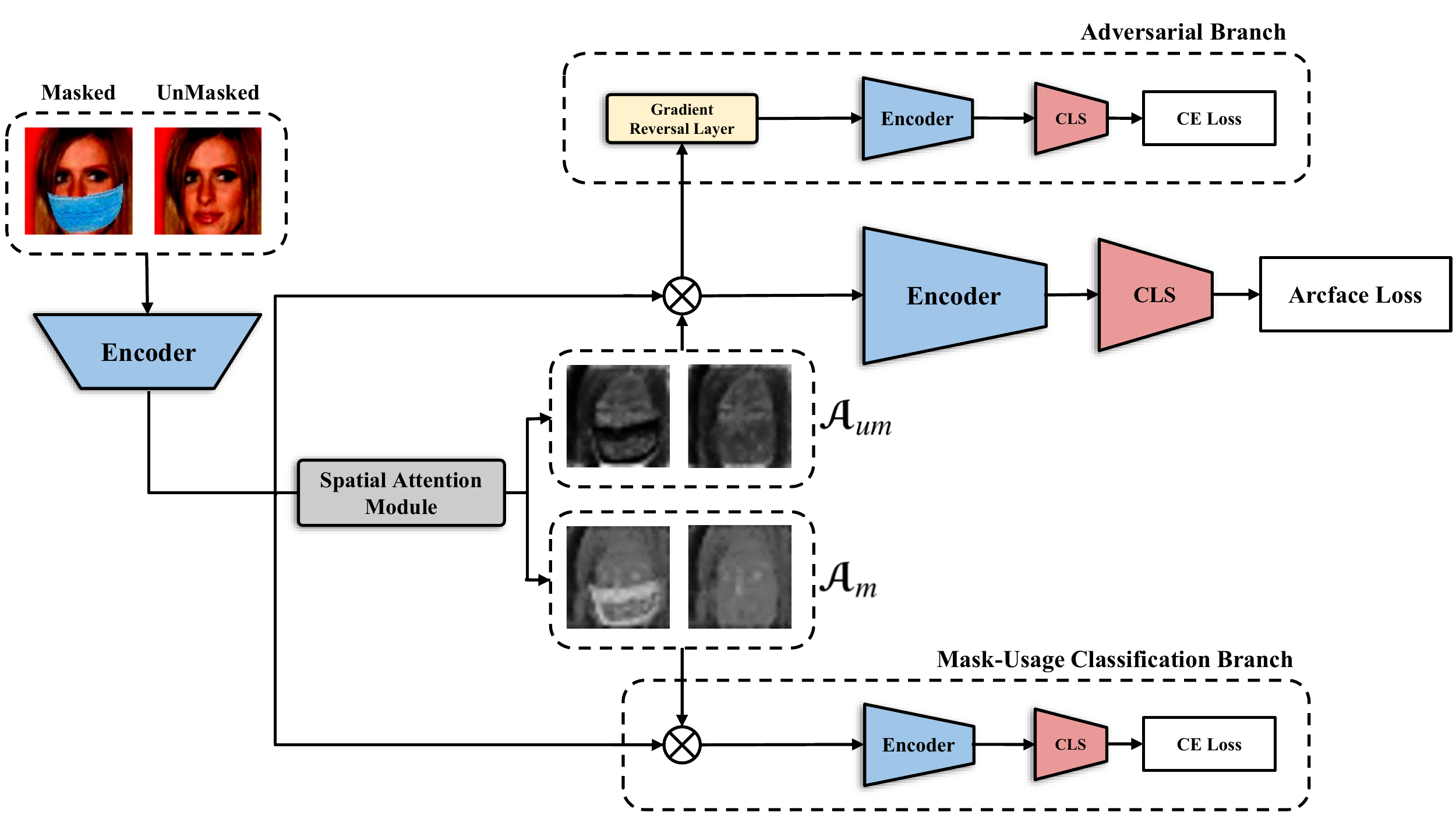}
    \caption{\textbf{Overview of proposed localization method.} Spatial attention module generates attention maps from input images to separate masked and unmasked regions. $A_{m}, A_{um}$ are masked attention and unmasked attention, respectively. To localize masked and unmasked regions without annotation, $A_m$ is used to learn face recognition with arcface loss and, $A_{um}$ is used to learn mask-usage classification with cross-entropy loss. We empirically verify that masked and unmasked regions are successfully localized, and the performances on both MFR and FR are improved with CAL}
    \vspace{-5mm}
    \label{fig:localizatoin}
\end{figure*}

\subsection{Complementary Attention Learning}
\label{subsec:comple}

To extract mask-invariant features while minimizing the loss of the FR performance, we propose the method that precisely separates a masked region by employing and enhancing the localization capability of the CBAM. First, we generate two attention maps, $\boldsymbol{\mathscr{A}_{um}}, \boldsymbol{\mathscr{A}_{m}}$ and two features, $\boldsymbol{X_{um}}, \boldsymbol{X_{m}}$, using the following equations:  
\begin{align}
    \boldsymbol{Z_s} &= \boldsymbol{ M_s( X_c )}, & \boldsymbol{\mathscr{A}_{um}} &= sigmoid(\boldsymbol{Z_s}), & \boldsymbol{\mathscr{A}_{m}} &= 1 -  \boldsymbol{\mathscr{A}_{um}},
\end{align}
\begin{align}
    \boldsymbol{X_{um}} &= \boldsymbol{\mathscr{A}_{um}} \otimes \boldsymbol{X_c}, & \boldsymbol{X_m} &= \boldsymbol{\mathscr{A}_{m}} \otimes  \boldsymbol{X_c},
\end{align}
where, $\boldsymbol{X_c}$ is the intermediate feature computed using eq~\ref{eq:cbam}, and $\boldsymbol{M_s}$ is the spatial attention module of the CBAM. We illustrate the procedure in the Fig~\ref{subfig:cbam}. $\boldsymbol{\mathscr{A}_{m}} \in \mathbb{R}^{1 \times H \times W} $ is the complementary attention of $\boldsymbol{\mathscr{A}_{um}} \in \mathbb{R}^{1 \times H \times W} $; hence, they activate mutually exclusively. Then, two distinct features $\boldsymbol{X_{um}}, \boldsymbol{X_{m}}$ are generated by mutually exclusive attention $\boldsymbol{\mathscr{A}_{um}}, \boldsymbol{\mathscr{A}_{m}}$, respectively. We use $\boldsymbol{X_{um}}$ for training face recognition, and it makes $\boldsymbol{\mathscr{A}_{um}}$ to localize the face region in an image. To prevent $\boldsymbol{\mathscr{A}_{um}}$ focus on the masked region of a face, we use $\boldsymbol{X_{m}}$ to train mask-usage classification. $\boldsymbol{X_{m}}$ is optimized to get information related to mask-usage. This makes $\boldsymbol{\mathscr{A}_{m}}$ focus on masked region, and prevents $\boldsymbol{\mathscr{A}_{um}}$ from being activated in the masked region owing to their complementary property. This method is named ``Complementary Attention Learning'' (CAL), and compare the performance of baseline with an auxiliary adversarial learning branch. We visualize the overall procedure of CAL and adversarial branch in the Fig~\ref{fig:localizatoin}. Adversarial branch is the same structure with the mask-usage branch, excepts there is the gradient reversal layer~\cite{ganin2015unsupervised} at the beginning of the branch. For comparison, we use $\boldsymbol{X_{um}}$ for training the adversarial branch, and it makes $\boldsymbol{X_{um}}$ invariant to mask-usage.

We train FR with arcface loss~\cite{deng2019arcface}, and train the mask-usage classification branch and adversarial branch with the cross-entropy loss~\cite{good1992rational}, as expressed by the following equations:

\begin{align}
    \mathscr{L}_{arc} = - \frac{1}{N_b} \sum^{N_{b}}_{i=1}log(\frac{e^{s(cos(\theta_{y_i} + m))}}{e^{s(cos(\theta_{y_i} + m))} + \sum^{N_c}_{j=1,j \neq y_i} e^{s(cos(\theta_j))}} ),
    \vspace{-2mm}
\end{align}

\begin{align}
    \theta_{j} = cos^{-1}( \frac{\boldsymbol{W_{j}} ^{\top} \cdot \boldsymbol{z_{i}}}{ \lVert \boldsymbol{W}_{j} \rVert  \lVert \boldsymbol{z_{i}} \rVert}  )
\end{align}

\begin{align}
    \mathscr{L}_{CE} = - \frac{1}{N_b} \sum^{N_b}_{i=1}log(\frac{e^{\boldsymbol{z_{y_i}}}}{\sum^{N_c}_{j=1} e^{\boldsymbol{z_{y_j}}}}),
\end{align}
where $N_b$ and $N_c$ are mini-batch size and the number of the classes, respectively. $y_i, \boldsymbol{z_i}$ is the target label index and logit of $\boldsymbol{x_i}$, respectively. $\boldsymbol{W} \in \mathbb{R}^{C \times N_c} $ is the weight of last linear classifier, where $C$ is the dimension size of the $\boldsymbol{z}$. $s$ and $m$ are the scale and margin of arcface loss, and $\theta_{j}$ is the angle between features $\boldsymbol{z_i}$ and weight $\boldsymbol{W_{j}}$. 

\begin{table*}[t]
    \begin{center}
        \begin{tabular} {l l c c c c c c}
            \toprule
            \multirow{2}{*}{Train Datasets} & \multirow{2}{*}{Models} & \multicolumn{6}{c}{Test Datasets}\\
            \cmidrule{3-8}
             & & MFR & MR-ALL & IJB-C & Age-DB & CALFW & CPLFW\\
            \toprule
            \multirow{16}{*}{CASIA} & Baseline &18.49 &23.65 & 69.63 & 94.07 &94.07 &88.97    \\
            &&&&&&\\
            & Baseline + MA=0.1 &31.23  & 26.35 &  \textbf{58.54} & \underline{93.96} & \underline{93.27} & 88.7 \\
            & Baseline + Adv + MA=0.1 & 33.36& 26.64  & 31.63 & 93.95 & 93.15 & 88.7 \\
            & CBAM + CAL + MA=0.1 & \underline{35.35}  & \textbf{28.58}  & \underline{50.36} & 93.92 & \textbf{93.35} & \textbf{89.2} \\
            & MFSA + CAL + MA=0.1 & \textbf{35.76} & \underline{27.61} & 48.21 & \textbf{94.03} & 93.22 & \underline{88.73}\\
            &&&&&&\\
            & Baseline + MA=0.3 & 40.37  & \underline{26.55} &  \textbf{47.82} & \underline{93.87} & \textbf{93.32} & \underline{88.82} \\
            & Baseline + Adv + MA=0.3 & 41.73& 24.31  & 30.62 & 93.33 & 93 & 88.37 \\
            & CBAM + CAL + MA=0.3 & \underline{42.1}  & 26.14  & 35.13 & 93.8 & 92.8 & 88.53 \\
            & MFSA + CAL + MA=0.3 & \textbf{43.44} & \textbf{28.94} & \underline{35.99} & \textbf{93.95}& \underline{93.07}&\textbf{88.92}\\
            &&&&&&\\
            & Baseline + MA=0.5 & 42.83  & 21.80 &  \textbf{18.34} & \textbf{93.07} & \underline{92.7} & 87.68 \\
            & Baseline + Adv + MA=0.5 & 43.15& \textbf{22.36}  & 8.92 & \underline{92.95} & 92.6 & \textbf{87.98} \\
            & CBAM + CAL + MA=0.5 & \underline{44.4}  & 20.74  & 11.45 & 92.68 & 92.52 & 87.9 \\
            & MFSA + CAL + MA=0.5 & \textbf{45.2} & \underline{21.87} & \underline{11.86} & 92.9&\textbf{92.83} &\underline{87.97}\\

            \midrule
            \multirow{6}{*}{MS1MV3} & Baseline & 65.86 & 80.53 & 94.80 & 98.30 & 96.17 & \textbf{92.90} \\
            &&&&&&\\
            & Baseline + MA=0.5 & 78.25  & \underline{69.41}  & \textbf{93.68} & \underline{97.90} & 96.03 & 92.50 \\
            & Baseline + Adv + MA=0.5 & \underline{78.48} & 68.71 & 93.54 & 97.03 & 95.13 & 92.30 \\
            & CBAM + CAL + MA=0.5 & 78.45  & 69.30  & 93.57 & \textbf{98.03} & \textbf{96.13} & 92.72 \\
            & MFSA + CAL + MA=0.5 & \textbf{78.70} & \textbf{69.64} & \underline{93.62} & \underline{97.90} & \underline{96.08}  & \underline{92.70}\\
            \bottomrule
        \end{tabular}
    \end{center}
    \vspace{-3mm}
    \caption{\textbf{Open-sourced face recognition datasets verification performances.} We report 1:1 verification TAR (@FAR=1e-5) on the IJB-C dataset, and verification performance (\%) of Age-DB, CALFW and CPLFW. ``MFR'' and ``MR-ALL'' denote TAR (@FAR=1e-4) on the masked test set and TAR (@FAR=1e-6) on the  multi-racial test set of the ICCV2021-MFR/Insightface track, respectively. ``MA'' means the masked face augmentation probability. Best in bold, second-best underlined.}
    \label{tab:main}
\end{table*}

\subsection{Multi-Focal Spatial Attention}

In the CBAM, the spatial attention module divides the region into binary regions, such the foreground and background. Then, with CAL, we train complementary attention $\mathscr{A}_{m}$ to localize a masked region and standard attention $\mathscr{A}_{um}$ to localize a unmasked region. In this case, it is ambiguous to classify a background region that does not belong to the masked and unmasked regions. Therefore, we propose the Multi-Focal Spatial Attention (MFSA) to apply N-way classification to a region. There are three classes for MFR (unmasked, masked, and background regions); hence, we use MFSA with 3-way classification. Then, MFSA can be expressed by the following formulas: 
\begin{align}
    \boldsymbol{Z_{s}} &= f(X), &  \boldsymbol{[\mathscr{A}_{um},\mathscr{A}_{m},\mathscr{A}_{bg}]} &= softmax(Z_s)
\end{align}
\begin{align}
    \boldsymbol{X_{um}} &= \boldsymbol{\mathscr{A}_{um}} \otimes \boldsymbol{X}, & \boldsymbol{X_m} &= \boldsymbol{\mathscr{A}_{m}} \otimes  \boldsymbol{X}, & \boldsymbol{X_{bg}} &= \boldsymbol{\mathscr{A}_{bg}} \otimes  \boldsymbol{X},
\end{align}
where $f(\cdot)$ is a network composed of pointwise convolution layers, batch normalization layers, and ReLU~\cite{ioffe2015batch, nair2010rectified}. To enhance the discriminative capability of $f(\cdot)$, we adopt Orthogonalization by Newton's Iteration (ONI)~\cite{huang2020controllable}. It orthogonalizes the weight of the pointwise convolution layers, and disentangle the attention representations. Orthogonalization is a popular technique, which is well-conditioning the network training behavior~\cite{cho2021improving, huang2019iterative}. Also, weight orthogonalization enforces each channel of $\boldsymbol{Z_{s}}$ to depend on a different input channel, so it reduces causality and correlation between attentions. To compute attention, $\boldsymbol{Z_{s}} \in \mathbb{R}^{3 \times H \times W}$ is normalized by $softmax$ function along the channel dimension. Finally, $\boldsymbol{\mathscr{A}_{um}},\boldsymbol{\mathscr{A}_{m}},\boldsymbol{\mathscr{A}_{bg}} \in \mathbb{R}^{1 \times H \times W}$ are element-wise multiplied to the input feature $\boldsymbol{X}$ to generate three features $\boldsymbol{X_{um}},\boldsymbol{X_{m}}, \boldsymbol{X_{bg}}$, respectively. We visualize the procedure of MFSA in the Fig~\ref{subfig:ours}. We utilize the $\boldsymbol{X_{um}}, \boldsymbol{X_{m}}$ by following the CAL described in Section~\ref{subsec:comple}. $\boldsymbol{X_{bg}}$ is not explicitly utilized during training, but it alleviates the ambiguity of the background region.

\begin{figure*}[!t]
    \centering
    \vspace{-6mm}
    \includegraphics[width=\linewidth]{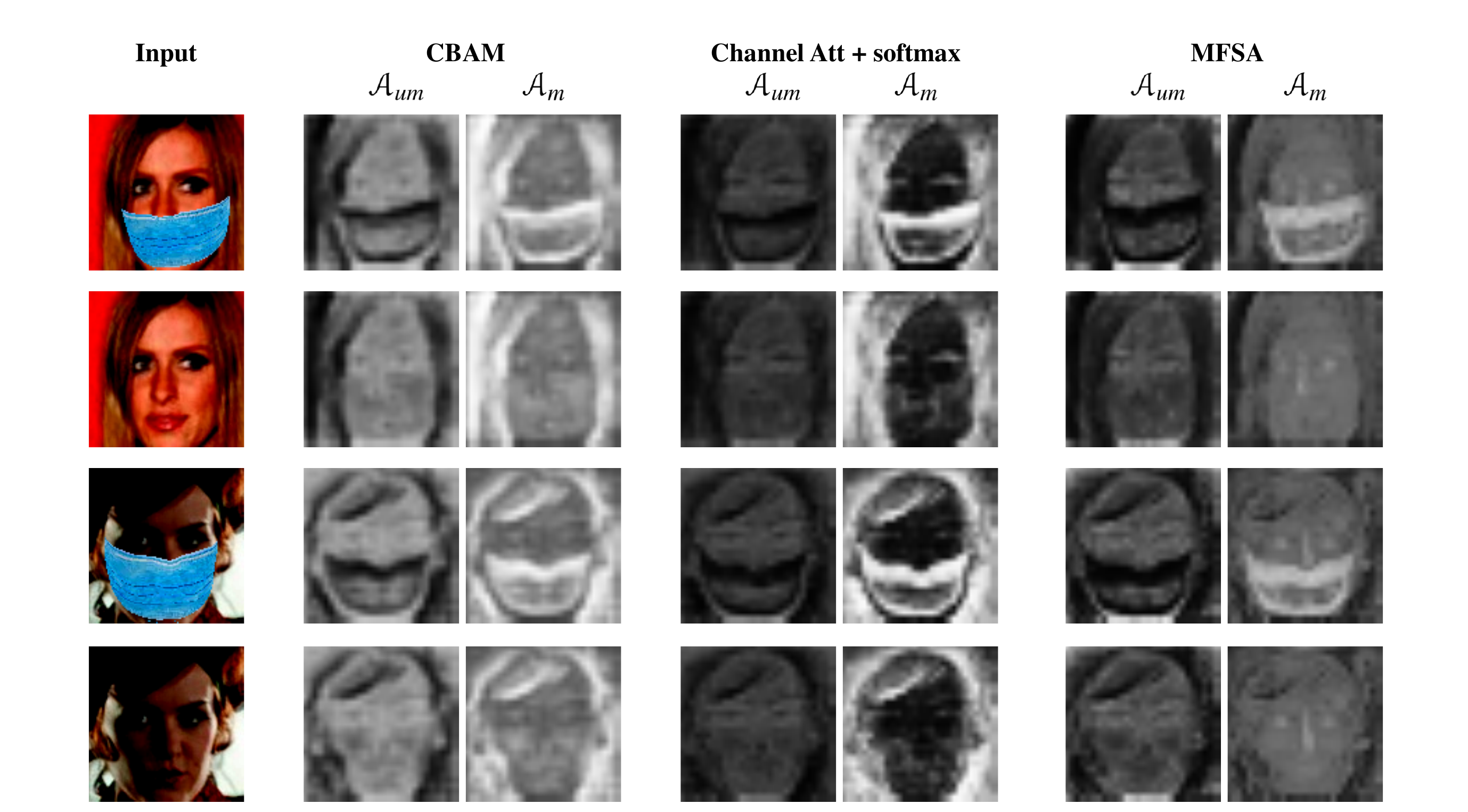}
    \caption{\textbf{Visualization of the attention maps.} From input images, spatial attention modules generates attention maps $\boldsymbol{\mathscr{A}_{um}}$, $\boldsymbol{\mathscr{A}_{m}}$. We visualize the attention maps as varying the attention modules. ``Channel Att $+$ $softmax$'' denotes the intermediate attention module that replace $sigmoid$ function of CBAM with $softmax$ function.}
    \label{fig:attention}
\end{figure*}

\section{Experiments}

\subsection{Training details}
We adopt ResNet-50~\cite{he2016deep} as the backbone architecture. We train networks using the standard data augmentation (i.e., flipping, translation, cropping), and mask augmentation using the tools introduced in the ICCV2021-MFR/Insightface track~\cite{deng2021masked}. We train the CASIA-Webface and MS1MV3 datasets~\cite{yi2014learning, deng2020retinaface, deng2019arcface, guo2016ms} by employing SGD with a mini-batch size of 512. Momentum and weight decay are set to 0.9 and 5e-4, respectively. We set initial learning rate to 0.2, and employ the polynomial learning rate decay scheduler~\cite{polyak1992acceleration, robbins1951stochastic} with 2 epochs of warm restart. We finish the training at 25 epochs and 34 epochs for MS1MV3 and CASIA-Webface datasets, respectively. Following the setup of~\cite{deng2019arcface}, we set the scale $s$ to 64 and the margin $m$ to 0.5 for arcface loss.

\subsection{Evaluation details}
We evaluate the conventional FR performance on the four benchmark FR datasets: IJB-C, Age-DB, CALFW, CPLFW~\cite{CPLFWTech, moschoglou2017agedb,DBLP:journals/corr/abs-1708-08197, maze2018iarpa}. Additionally, we report the performance on the multi-racial face dataset (MR-ALL) and masked face dataset (MFR) provided by the ICCV2021-MFR/Insightface track~\cite{deng2021masked}. 

\noindent \textbf{Masked Test Set of ICCV2021-MFR/Insightface track (MFR):} It contains 6,964 real-world masked facial images and 13,928 unmasked facial images. There are 20,892 images of 6,964 identities. We evaluate the 1:1 face verification TAR (@FAR=1e-4). 

\noindent \textbf{Multi-Racial Test Set of ICCV2021-MFR/Insightface track (MR-ALL):} It consists of four demographic groups (African, Caucasian, South Asia, and East Asia), and contains 1.6M images of 242K identities. We evaluate the 1:1 face verification TAR (@FAR=1e-6). 

\noindent \textbf{IJB-C:} It is large-scale face recognition dataset that contains 148.8K images of 3,531 identities. We evaluate the 1:1 face verification TAR (@FAR=1e-5). 

To investigate efficacy of proposed method, we evaluate the performance of FR and MFR as varying the spatial attention modules (CBAM, MFSA) and training methods (Adv, CAL). ``Baseline'' denotes the performance of ResNet-50 trained with arcface loss. ``MA'' means the probability of the masked face augmentation. ``Adv'' denotes the adversarial learning with auxiliary adversarial branch to make features invariant to mask-usage.

From the results shown in Table~\ref{tab:main}, we verify that Multi-Focal Spatial Attention (MFSA) with Complementary Attention Learning (CAL) demonstrates the best performance on MFR regardless of the MA probability and train datasets. Additionally, CAL shows smaller IJB-C performance drop than for the Adv. For IJB-C, Baseline$+$MA obtains the best performance, but CAL$+$MA obtains the second best performance. It indicates CAL successfully enhances the localization capability of spatial attention modules and more precisely extracts mask-invariant features. Therefore, the performances of conventional FR datasets are less degenerated.

\subsection{Qualitative Results}
We visualize two attention maps $\boldsymbol{\mathscr{A}_{um}}$ and $\boldsymbol{\mathscr{A}_{m}}$ as varying the spatial attention modules in the Fig~\ref{fig:attention}. As we expected, in the CBAM, background regions are ambiguous to be classified. Therefore, background regions are not clearly removed on $\boldsymbol{\mathscr{A}_{um}}$, and it is not desirable phenomenon. ``Channel Att $+$ $softmax$'' denotes the attention module that replace $sigmoid$ function of CBAM with $softmax$ function. In the ``Channel Att $+$ $softmax$'', the background regions are not ambiguous to classified, but unmasked regions are not activated at $\boldsymbol{\mathscr{A}_{um}}$. Also, background regions are activated at $\boldsymbol{\mathscr{A}_{m}}$. We suspect that the representation capacity should be increased to alleviate the problem. Therefore, we propose MFSA that replace channel attention with convolution layers with ONI, and obtain the desirable results. $\boldsymbol{\mathscr{A}_{um}}$ of MFSA is relatively invariant to the mask-usage, and unmasked, masked, and background regions are more clearly divided.

\subsection{Ablations}
\begin{table}[ht]
\begin{center}
    \begin{tabular}{l c c c}
    \toprule
    Models & MFR & MR-ALL & IJB-C\\
    \toprule
    Baseline &18.49 &23.65 & \textbf{69.63}   \\
    Baseline + MA=0.1 &\textbf{31.23}  & \textbf{26.35} &  58.54\\
    &&&\\
    Baseline + Adv + MA=0.1 & 33.36& 26.64  & 31.63  \\
    CBAM + MA=0.1 & 32.71& 27.24  & 37.20  \\
    CBAM + Adv + MA=0.1 & 34.04 &27.47 & 33.66   \\
    CBAM + CAL + Adv + MA=0.1 & 33.89  &27.148  &40.02  \\
    CBAM + CAL + MA=0.1 & \textbf{35.35}  & \textbf{28.5}8  & \textbf{50.36} \\
    \bottomrule
    \end{tabular}
\end{center}
\vspace{-3mm}
\caption{Comparisons of 1:1 verification performance ($\%$) on MFR, MR-ALL, and IJB-C. CAL shows the best performance on all test sets. Best in bold.}
\label{tab:abl}
\end{table} 
To investigate the efficacy of CAL, we conduct ablation studies as varying the training methods. As shown in Table~\ref{tab:abl}. We verify that MA improves the MFR performance, but degenerates the MR-ALL and IJB-C performance. With Adv, the performance of MFR is improved by $2.13\%$, but the performance of IJB-C is significantly decreased by $16.91\%$. By contrast, CBAM with CAL improves the MFR performance by $3.12\%$, and decreased the FR performance by $8.18\%$. Notably CAL $+$ Adv shows the worse performance than CAL. It indicates that CAL is more effective training method to extract mask-invariant features than adversarial learning.

\section{Conclusion}
In this paper, we propose the method to extract the mask-invariant features by employing and enhancing the localization capability of the CBAM, which is a representative attention module. First, we propose the Complementary Attention Learning (CAL) that adversarially utilizes the complementary attention to prevent the standard attention is being activated on the undesirable area. From the ablation studies, we empirically demonstrate that CAL is more efficient to extract mask-invariant feature than simple adversarial learning. Second, we propose the Multi-Focal Spatial Attention (MFSA) that divides a image into N-way. It alleviates the ambiguity of the background regions classification. From the visualized attention maps of CBAM and MFSA, we verify that MFSA successfully neglects the background regions and extracts mask-invariant features. Additionally, MFSA with CAL gets the best MFR performance regardless of the train dataset and MA probability with relatively smaller FR performance degeneration.

{\small
\bibliographystyle{ieee}
\bibliography{egbib}
}

\end{document}